\documentclass[lettersize,journal]{IEEEtran}
\usepackage{amsmath,amsfonts}
\usepackage{algorithmic}
\usepackage{algorithm}
\usepackage{array}
\usepackage[caption=false,font=normalsize,labelfont=sf,textfont=sf]{subfig}
\usepackage{textcomp}
\usepackage{stfloats}
\usepackage{url}
\usepackage{verbatim}
\usepackage{graphicx}
\usepackage{cite}
\usepackage{bm}
\usepackage{algorithm}
\usepackage{mathtools}
\usepackage[hidelinks,colorlinks=false]{hyperref}

\begin{document}

\title{LCAMV: High-Accuracy 3D Reconstruction of Color-Varying Objects Using LCA Correction and Minimum-Variance Fusion in Structured Light}

\author{Wonbeen Oh, and Jae-Sang Hyun,~\IEEEmembership{Member,~IEEE}
        % <-this % stops a space
\thanks{Corresponding author: Jae-Sang Hyun}
\thanks{Wonbeen Oh and Jae-Sang Hyun are with the Department of Mechanical Engineering, Yonsei University, Seoul 03722, South Korea. (e-mail: hyun.jaesang@yonsei.ac.kr)}}

\markboth{Journal of \LaTeX\ Class Files,~Vol.~14, No.~8, August~2021}%
{Shell \MakeLowercase{\textit{et al.}}: A Sample Article Using IEEEtran.cls for IEEE Journals}

\IEEEpubid{0000--0000/00\$00.00~\copyright~2021 IEEE}

\maketitle

\begin{abstract}
Accurate 3D reconstruction of colored objects with structured light (SL) is hindered by lateral chromatic aberration (LCA) in optical components and uneven noise characteristics across RGB channels. This paper introduces lateral chromatic aberration correction and minimum-variance fusion (LCAMV), a robust 3D reconstruction method that operates with a single projector–camera pair without additional hardware or acquisition constraints. LCAMV analytically models and pixel-wise compensates LCA in both the projector and camera, then adaptively fuses multi-channel phase data using a Poisson–Gaussian noise model and minimum-variance estimation. Unlike existing methods that require extra hardware or multiple exposures, LCAMV enables fast acquisition. Experiments on planar and non-planar colored surfaces show that LCAMV outperforms grayscale conversion and conventional channel-weighting, reducing depth error by up to 43.6\%. These results establish LCAMV as an effective solution for high-precision 3D reconstruction of nonuniformly colored objects.
\end{abstract}


\section{Introduction}
\label{sect:Introduction}

\IEEEPARstart{3}{D} reconstruction using structured light (SL) has been extensively researched for the past 20 years. Its high precision has enabled applications in manufacturing \cite{wang2021development}, medical imaging \cite{olesen2011motion}, and robotics \cite{yang2020advances}. Recent advances in graphics processing units (GPUs) have expanded the use of 3D geometry to emerging fields such as virtual reality \cite{anthes2016state} and teleoperation \cite{zhang2018deep}.
The rise of computer vision algorithms incorporating artificial intelligence has also unlocked new applications of 3D geometry data, such as spatial AI \cite{mildenhall2021nerf,kerbl20233dgaussiansplattingrealtime} and robotic AI \cite{goyal2023rvt}.
As these applications increasingly require capturing real-world objects with both precise geometry and accurate color information, the ability to extract reliable 3D geometry together with true RGB data has become critical. However, achieving such accuracy is particularly difficult for objects with nonuniform surface colors, as color-dependent responses often degrade reconstruction quality.

\begin{figure}[!t]
\centering
\includegraphics[width=88mm]{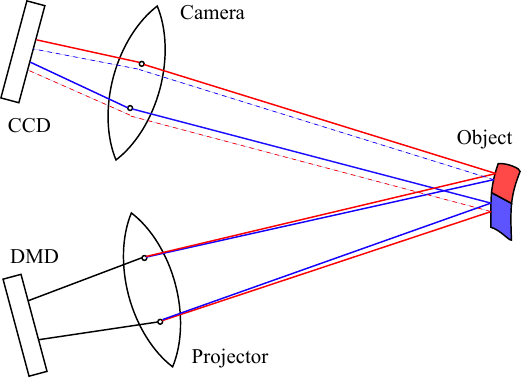}
\caption{LCA on scanning a colored object in DFP system.}
\label{fig:problem1}
\end{figure}

The first challenge is \textbf{lateral chromatic aberration (LCA)}. Because the refractive index of optical lenses varies with wavelength, Because of the refractive index of a lens with wavelength, rays of different colors are refracted by different amounts, affecting both the projector and camera. In a typical DMD-based projector, white light is sequentially generated by rapidly displaying red, green, and blue channels. Each color channel follows a slightly different optical path because of LCA in the projector lens, resulting in their projections being spatially shifted on the object surface; similarly, the camera lens introduces additional LCA when capturing these channels. Prior studies \cite{chen2020correcting, huang2016analysis, chen2023flexible, sun2021analysis} on LCA in digital fringe projection have primarily addressed systems using composite color-fringe patterns, where LCA directly degrades phase-shifting accuracy. However, we emphasize that the LCA is also a critical source of error when scanning nonuniformly colored objects using nomially white fringe patterns. For example, consider scanning a surface containing distinct red and blue regions. In the captured fringe images, the red and blue regions are predominantly illuminated by the projector's respective red and blue channels, which undergo different refractions due to LCA, as illustrated in Fig. \ref{fig:problem1}.

\IEEEpubidadjcol % To avoid Collision
Beyond LCA, another fundamental difficulty arises from \textbf{the unequal sensitivity across the RGB channels}. For a given pixel with channel-wise RGB intensity values, each channel exhibits different sensitivity, depending on the object's local color. For example, when scanning a red object, the captured image contains the most signal in the red channel, less in the green channel, and predominantly noise in the blue channel. Taking mean intensity across RGB or using predefined weights for RGB to grayscale, such as those Y'UV color model \cite{macadam1937projective}, is not optimal solution for achieving high reconstruction precision.

\subsection{Related Works}

Various approaches have been proposed to simultaneously acquire 3D geometry and texture of colored objects using digital fringe projection. Zhang \cite{zhang2004high} employed a beam splitter to allow a color camera and a monochrome camera to capture the same scene simultaneously; the monochrome images were used to reconstruct the 3D geometry while the color images provided the RGB texture. In subsequent work, Zhang \cite{zhang2008simultaneous} utilized a single RGB camera by performing phase unwrapping directly on the raw Bayer image and applying bilinear interpolation to obtain the RGB texture. Although this approach reduced the quality of the reconstructed 3D geometry, it enabled simultaneous acquisition of geometry and RGB texture using a single camera. Other studies \cite{ou2013flexible, xu2014simultaneously} improved 3D geometry quality by utilizing infrared light, which does not interfere with visible illumination. However, these methods require additional hardware—such as beam splitters, IR light sources, or extra cameras—to achieve accurate simultaneous RGB-D reconstruction. Few studies have rigorously investigated why using RGB images for 3D reconstruction often leads to reduced accuracy. Recently, Wang \cite{wang2020rapid} applied high dynamic range (HDR) imaging with exposure-time adjustment during RGB light projection to mitigate varying signal-to-noise ratios (SNR) caused by nonuniform surface reflectance. However, this method requires multiple captures with varying exposure times and color projections. Yuan \cite{yuan2020absolute} proposed robust scanning of colored objects using a specialized 3-PSA code. This method focuses on robustly determining the fringe order for phase unwrapping when scanning colored objects with fewer patterns, rather than minimizing error and improving 3D geometry quality within phase-shifting algorithms Although both approaches demonstrated robust performance when scanning colored objects, they did not account for LCA effects from either the camera or the projector.

\subsection{Contribution}

We propose a novel method, LCAMV (lateral chromatic aberration correction and minimum-variance channel fusion), which robustly reconstructs 3D geometry of nonuniformly colored objects using a single RGB camera and a projector pair, requiring no additional hardware. Furthermore, the method is fully software-based and requires \textbf{no additional image captures}, making it potentially applicable to high-speed 3D reconstruction. Our method consists of two stages: 1) Channel-wise LCA correction: Correcting the projector and camera's LCA to align information from the RGB channel properly. 2) Minimum variance channel fusion: Merge aligned RGB information adaptively to maximize the 3D geometry's precision. To the best of our knowledge, this is the first work to explicitly address the LCA in DFP when using nominally white fringe patterns on colored objects.

In this paper, we first analyze the theoretical impact of projector-induced LCA and channel-dependent RGB noise captured by the camera in Section~\ref{sect:Preliminary}. We then present our method, which addresses both issues using optical modeling and statistical theory, in Section~\ref{sect:Method}. Finally, we validate the proposed approach through real-world experiments in Section~\ref{sect:Experiment}.

\section{Principles}
\label{sect:Preliminary}

\begin{figure}[t]
  \centering
  \includegraphics[width=88mm]{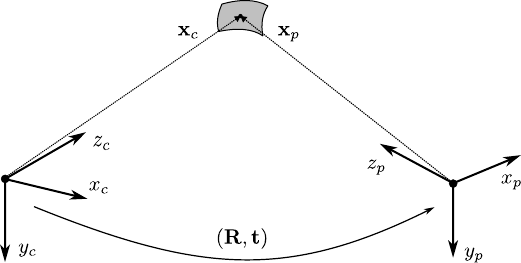}
  \caption{Structured light system setup}
  \label{fig:SLS}
\end{figure}

\noindent 3D geometry of an object can be captured with a camera–projector pair in typical DFP. Similar to 3D reconstruction using a stereo-vision system \cite{ma2004invitation}, we can consider the projector as an inverse pinhole camera model and conduct \textit{triangulation} as shown in \eqref{eqn:triangulation}. In detail, the geometric relation between the camera and the projector image plane can be represented by the following equation:
\begin{subequations}\label{eqn:triangulation}
\begin{align}
  \mathbf{x}_{p} &= \mathbf{R}\mathbf{x}_{c} + \mathbf{t} \\
  \mathbf{x}_{c} &= s_{c}\mathbf{K}_c^{-1}\mathbf{u}_c \\
  \mathbf{x}_{p} &= s_{p}\mathbf{K}_p^{-1}\mathbf{u}_p
\end{align}
\end{subequations}

where $\mathbf{x}_c, \mathbf{x}_p \in \mathbb{R}^3$ are 3D vectors of a point in the camera and projector coordinate systems, and $\mathbf{u}_c, \mathbf{u}_p \in \mathbb{R}^3$ are in pixel coordinates. Note that pixel vectors $\mathbf{u}$ are in homogeneous coordinates:$\mathbf{u}_c=[u_c,v_c,1]^T$. $s_c, s_p$ are scalar values representing the depth from the camera and projector. $\mathbf{K}_c, \mathbf{K}_p \in \mathbb{R}^{3\times3}$ are three-by-three intrinsic matrices which map camera and projector coordinates to pixel coordinates, $\mathbf{R} \in SO(3)$ denotes the rotational matrix, which maps projector to camera coordinates, and $\mathbf{t}\in \mathbb{R}^{3}$ denotes the translation vector from projector to camera coordinates, as shown in Fig. \ref{fig:SLS}.

Note that $\mathbf{K}_c, \mathbf{K}_p, \mathbf{R}, \mathbf{t}$ are predefined by camera and projector calibration. Given a certain camera pixel $(u_c,v_c)$, there are four unknowns $u_p, v_p, s_p, s_c$ with three equations. Thus, we can calculate for depth from camera $s_c$ and projector $s_p$ if the projector pixel coordinate is given. In this paper, we will consider the acquisition of the projector's horizontal pixel $u_p$ for 3D geometry reconstruction. Assuming $u_p$ is known, $s_c$ and $s_p$ can be calculated by using the first and third rows of the following equations:
\begin{equation}
  s_p\mathbf{u}_p=s_c\mathbf{K}_p\mathbf{R}\mathbf{K}_c^{-1}\mathbf{{u}}_c+\mathbf{K}_p\mathbf{t}
  \label{eqn:triangulation_detail}
\end{equation}

which can be derived from \eqref{eqn:triangulation}. Furthermore, we can solve for the projector's vertical pixel $v_p$ using epipolar constraints:
\begin{equation}
  \mathbf{u}_p^T\mathbf{F}\mathbf{u}_c=0
  \label{eqn:fundamentalmatrix}
\end{equation}

where $\mathbf{F}=\mathbf{K}_p^{-T}[\mathbf{t}]_{\times}\mathbf{R}\mathbf{K}_c^{-1}$ is a fundamental matrix which can be calculated with calibrated components, and $[\cdot]_{\times}$ is a cross product matrix. $v_p$ will be further used for the LCA correction algorithm in Section \ref{subsect:LCACorrection}. Borrowing the concept of the \textit{phase shifting} from interferometry \cite{creath1988v}, $u_p$ can be acquired. The projector pixel $u_p$ is scaled to phase $\phi$ by the following equation:
\begin{equation}
    \phi=(2\pi /\lambda)\cdot u_p,\quad \phi\in[0,2L\pi)
\end{equation} 

where $\lambda$ is the arbitrarily chosen wavelength of fringe patterns in pixels, and $L$ is the number of phase periods. Note that the $L\cdot\lambda>W_p$ condition should be met, where $W_p$ is the projector pixel width, to fully cover the projector's pixel domain. By capturing the scene upon projection of phase-modulated fringe patterns, we can find corresponding $\phi$ in the camera image. For the typical three-step phase shifting algorithm \cite{creath1988v}, the intensity of three captured images in the camera has the following relation:
\begin{subequations}\label{eqn:3stepshift}
\begin{align}
  I_1[u_c,v_c] &= I_A + I_B \cos\left(\phi - \frac{2}{3}\pi\right) \\
  I_2[u_c,v_c] &= I_A + I_B \cos(\phi) \\
  I_3[u_c,v_c] &= I_A + I_B \cos\left(\phi + \frac{2}{3}\pi\right)
\end{align}
\end{subequations}

where $I_1, I_2, I_3$ are the captured intensities at the $\left(u_c,v_c\right)$ pixel, $I_A$ is the average intensity, and $I_B$ is the modulated intensity by fringe patterns. Note that we can solve for $I_A,I_B$, and reconstruct the texture of the image upon uniform light projection by adding $I_A$ and $I_B$ \cite{zhang2008simultaneous}. By the three-step phase-shifting algorithm, the wrapped phase $\phi_w$ can be acquired as follows:
\begin{equation}
  \phi_w[u_c,v_c] =\tan^{-1}\left({\frac{\sqrt{3}(I_1-I_3)}{2I_2-I_1-I_3}}\right)
  \label{eqn:arctan_phaseshift}
\end{equation}

Note that the wrapped phase ranges $\phi_w \in \left[-\pi,+\pi\right)$, thus \textit{phase unwrapping} is required to fully reconstruct original $\phi$. Phase unwrapping can be done by acquiring fringe order $l\in \{0, \ldots, L-1\}$ using binary gray-code decoding \cite{sansoni1999three} or other phase unwrapping algorithms \cite{zhang2018absolute, zuo2016temporal}. Then, the original phase can be reconstructed as follows:
\begin{equation}
  \phi = \phi_w +2\pi l + \pi
  \label{eqn:phaseunwrap}
\end{equation}

Given reconstructed $\phi$, we can linearly scale into $u_p$, then solve for \eqref{eqn:triangulation_detail} to get 3D geometry. Note that multiple number of steps can also be used, by generalized N-step phase shifting algorithm solving least squares \cite{bruning1974digital, greivenkamp1984generalized}. 

\begin{figure}[t]
  \centering
  \includegraphics[width=88mm]{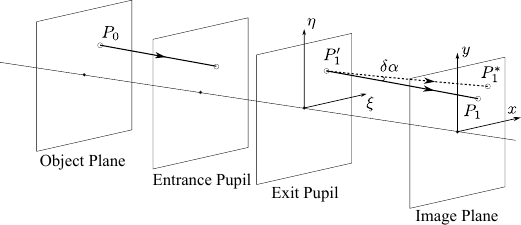}
  \caption{Aberration in camera system}
  \label{fig:cam_LCA}
\end{figure}

\subsection{Lateral Chromatic Aberration in Camera and Projector}

Chromatic aberration (CA) appears in optical systems capturing multiple wavelengths of light. A lens made of glass has a different refractive index with respect to wavelength, which makes light of different colors be projected onto different locations. If this difference is observed parallel to the optical axis, it is called axial chromatic aberration (ACA), and if orthogonal to the optical axis, lateral chromatic aberration (LCA). ACA alters the effective focal length, leading to different blur levels across wavelengths, while LCA shifts the actual projection location on the image plane. Since difference in projection location leads to triangulation error, LCA becomes more problematic when it comes to accurate 3D reconstruction.

Chromatic aberration can be modeled using the \textit{wavefront aberration function} \cite{born2013principles}, which models the wave aberration $\Psi$ between ideal and real-world optics. In detail, wave aberration can be viewed as a wavefront error, which is an optical path length difference between an ideal spherical wavefront and an actual wavefront. Given wave aberration $\Psi$, ray aberration $\delta\alpha$, and the resulting lateral aberration at the image plane $\Delta=P_1-P_1^*$, they can be expressed as follows:
\begin{subequations}\label{eqn:waveaberration}
\begin{align}
  & \delta\alpha_{x} = \frac{\partial \Psi}{\partial \xi}, \quad 
      \delta\alpha_{y} = \frac{\partial \Psi}{\partial \eta} \\
  & \Delta_x = -\frac{D}{n\gamma} \delta\alpha_x, \quad 
      \Delta_y = -\frac{D}{n\gamma} \delta\alpha_y
\end{align}
\end{subequations}

where $(x,y)$ are image coordinates and $(\xi,\eta)$ are pupil coordinates, $D$ is a distance between the exit pupil and the image plane, $n$ is a refractive index of the medium, and $\gamma$ is a scaling coefficient from $(\xi,\eta)$ to $(x,y)$, as shown in Fig. \ref{fig:cam_LCA}.

Under the rotational symmetric assumption, aberration can be approximated using the fourth-order expansion of $\Psi$, from \textit{Seidel eikonal perturbation} \cite{born2013principles}. Considering only the distortion term of the wave aberration function leads to:
\begin{equation}
  \Psi(x,y;\xi,\eta)=c(x^2+y^2)(x\xi+y\eta)
  \label{eqn:aberration_distortion}
\end{equation}

where $c$ is a coefficient. Then, we can consider decentering effects in the image plane using perturbations: $x\rightarrow x+\epsilon_x$, $y\rightarrow y+\epsilon_y$. Neglecting $\epsilon^2$ terms, \eqref{eqn:aberration_distortion} leads to:
\begin{multline}
  \Psi(\cdot)=cr^2\kappa^2+c\xi(\epsilon_x(3x^2+y^2)+2\epsilon_yxy)+\\
  c\eta(2\epsilon_xxy+\epsilon_y(3y^2+x^2))
\end{multline}

where $r^2=x^2+y^2$ and $\kappa^2=x\xi+y\eta$. Then, we can plug $\Psi$ into \eqref{eqn:waveaberration} to find out the lateral aberration at the image plane. For considering chromatic aberration, the contribution of different refractive indices with respect to wavelengths can be done by adding an extra linear term \cite{mallon2007calibration}. Then, the equation for the LCA model leads to:
\begin{align}
\begin{aligned}
  \Delta_{x}^{g}(\mathbf{x}_f,\mathbf{c}_g) =c_{1}x_{f}+c_{2}x_{f}r_{f}^{2}+c_{3}(3x_{f}^{2}+y_{f}^{2})+2c_{4}x_{f}y_{f} \\
  \Delta_{y}^{g}(\mathbf{x}_f,\mathbf{c}_g)=c_{1}y_{f}+c_{2}y_{f}r_{f}^{2}+2c_{3}x_{f}y_{f}+c_{4}(3y_{f}^{2}+x_{f}^{2})
\end{aligned}
\label{eqn:LCApolynomial}
\end{align}

where $\mathbf{x}_f=[x_{f},y_{f}]$ is an image coordinate vector of the reference wavelength $f$, and $\mathbf{c}_g=[c_1,c_2,c_3,c_4]$ is a coefficient vector of the specific wavelength $g$, which has to be corrected. Note that the optical system, including multiple rotational-symmetric components, can be modeled as linear combinations of the same equation \cite{born2013principles}; thus, multiple components introduce no additional parameters.

\begin{figure}[t]
  \centering
  \includegraphics[width=88mm]{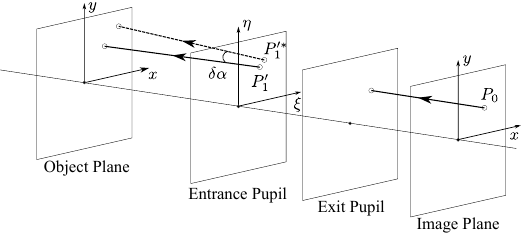}
  \caption{Aberration in projector system}
  \label{fig:prj_LCA}
\end{figure}

Unlike camera systems where the distance between the image sensor and exit pupil is constrained, the distance between the projector's entrance pupil and the object is not constrained and is comparatively large. Therefore, error induced by the projector's LCA is much larger than the camera's LCA \cite{sun2021analysis,huang2016analysis}. Inspired by prior works modeling aberration in camera \cite{mallon2007calibration} and projector \cite{sun2021analysis}, we will model the projector's LCA, treating a projector as an inverse camera model. In the projector system, we have to consider both ray aberration $\delta\alpha$ and lateral aberration $\Delta_E=P_1'-P_1'^*$ at the entrance pupil for optical systems with multiple components, as shown in Fig. \ref{fig:prj_LCA}. 

In the camera system it was possible to treat LCA using a depth-independent model since distances between every plane are uniform across all pixels. However, in the projector system, the distance to the object depth $z$ is arbitrary. Therefore, the resulting lateral aberration $\Delta_{O}=P_1-P_1^*$ at the object will be a function of $z$, which can be modeled by a linear model consisting of two parameters $\alpha$ and $\beta$ as follows. 
\begin{equation}
  \Delta_{O}=\alpha \cdot z+\beta
\label{eqn:projector_aberration}
\end{equation}

where the first-order term is equal to the ray aberration $\alpha=\delta\alpha$, and the constant term represents the distance between the intersections of the two lines, $\overline{P_1^*P_1'^*}$ and $\overline{P_1P_1'}$, with the image plane at $z=0$. Note that $z$ is a depth from the projector, which is contradictory in that we need ground truth depth $z$ to compensate for the $\Delta_O$. We use \textit{plug-in error estimation} \cite{carroll2006measurement}, approximating the depth-dependent $\Delta_O$ using the corrupted observation $z'$ under small error assumption. The error between the ground truth $\Delta_O(z)$ and approximated $\Delta_O(z')$ can be expressed as follows:
\begin{equation}
  \Delta_O(z)-\Delta_O(z')=\alpha(z-z')
\end{equation}

Since the depth difference induced by the LCA effect is very small $(z\approx z')$ and the ray aberration coefficient is also small $\alpha \ll1$, the lateral aberration can be approximated using the initial depth estimate $z'$. Since we do not have access to the ground-truth depth $z$, and for notational simplicity, we will use $z$ to denote the estimated depth $z'$ from this point.

Unfortunately, off-the-shelf projectors, which can project arbitrary RGB images, have complicated optical components such as dichroic prisms and mirrors \cite{TI_DLPA078C}; thus, its LCA is not guaranteed to be modeled using \eqref{eqn:LCApolynomial} under the rotational-symmetric assumption. To solve this, we used the Look-Up-Table (LUT) approach by actually finding out data points of $\Delta_O$ and $z$ for every projector pixel by scanning the white plate.

\begin{figure}[t]
  \centering
  \includegraphics[width=88mm]{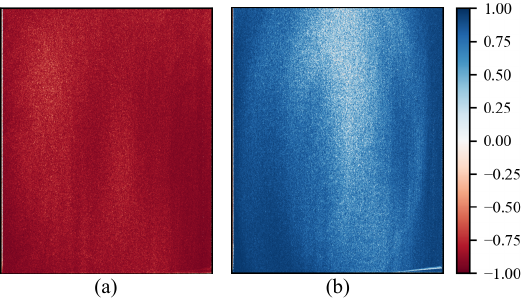}
  \caption{Pixel-wise Pearson correlation map of $\Delta_O$ and $z$ from channel (a) red (b) blue}
  \label{fig:correlation}
\end{figure}

Fig. \ref{fig:correlation} is a projector pixel-wise Pearson correlation coefficient \cite{leon2017probability} map $\rho[u_p,v_p]$ between observed lateral aberration $\Delta_O$ and depth $z$ by scanning the white plate. A total of $N=18$ samples were used from depths ranging in approximately $[180,340]$ millimeters. It can be found that except for the center-top region of the blue channel where the LCA effect is comparatively low, $\Delta_O$ and $z$ are highly linearly correlated with the opposite sign, which shows the validity of our model.

\subsection{Noise Analysis of Intensity and Phase}

Due to the nonuniform color setup, some color channels will contain information while others are relatively noisy, depending on the object's surface reflectance. For the noise analysis, we will assume the noise of RGB channels as a \textit{Poisson-Gaussian} model \cite{healey2002radiometric,foi2008practical} as follows:
\begin{subequations}\label{eqn:poissongaussian}
\begin{align}
  I[u_c,v_c] &= \mu_{I}[u_c,v_c] + n[u_c,v_c] \\ 
  n &\sim \mathcal{N}(0, \sigma_n^2(I)) \\
  \sigma_n^2(I) &= k_0 + k_1 \cdot I
\end{align}
\end{subequations}

where $\mu_{I}$ is an ideal ground truth intensity, $I$ is an observed noise-corrupted intensity, and $n$ is an approximated gaussian noise, with variance of $\sigma_n^2$ at the $(u_c,v_c)$ pixel. The variance of the noise is approximated with a linear model, where coefficient $k_0$ models a signal-independent read noise and $k_1$ models a signal-dependent shot noise of the captured image. Note that due to different optical properties of the Bayer color filter and different gain across RGB channels, noise parameters $k_0$ and $k_1$ would be different with respect to RGB channels.

Recall that 3D geometry reconstruction is done by triangulation using a delegate projector pixel, $\phi$. To find the contribution of the intensity and its noise to the acquired phase $\phi$, we should model the uncertainty of phase $\sigma_\phi^2$, deriving from phase shifting algorithms. According to \cite{wang2022phase}, under the constant Gaussian noise assumption, the variance of the wrapped phase error using the generalized N-step phase shifting algorithm can be expressed as:
\begin{equation}
  \sigma_{\phi_w}^2\approx\frac{2\sigma^2}{NI_B^2}
  \label{eqn:phasevariance}
\end{equation}

where $\sigma^2$ is the variance of intensity noise. In our Poisson-Gaussian model, the variance of the noise is a function of intensity as shown in \eqref{eqn:poissongaussian}. From \eqref{eqn:phasevariance}, we replace the constant variance $\sigma^2$ using the mean noise level $\bar{\sigma}_{n}^2$ across intensities $I_i$. Since $\sigma_{n_i}^2$ is linearly proportional to intensity, mean noise level is simply the noise of the mean intensity $I_A$. Then, \eqref{eqn:phasevariance} under the Poisson-Gaussian model leads to:
\begin{equation}
  \sigma_{\phi_w}^2\approx\frac{2(k_0+k_1I_A)}{NI_B^2}
  \label{eqn:phasevariance_PoissonGaussian}
\end{equation}

where $I_A$ is the mean intensity and $I_B$ is the modulation intensity. For the exemplary three-step phase-shifting algorithm, $I_A,I_B$ can be calculated as follows:
\begin{subequations}
\begin{align}
  I_A&=(I_1+I_2+I_3)/3 \\
  I_B&=\sqrt{3(I_1-I_3)^2+(2I_2-I_1-I_3)^2}/3
\end{align}
\end{subequations}

We can consider the numerator term as the mean noise $\bar{\sigma}^2_{n}=k_0+k_1I_A$, and the denominator term is the square of the modulated signal $I_B$ due to fringe image projections. Then, the uncertainty of phase $\sigma_{\phi_w}^2$ is inversely proportional to SNR. In this paper, we will consider the typical binary gray-code decoding \cite{sansoni1999three} for the phase-unwrapping algorithm; thus, the uncertainty of phase is the uncertainty of the wrapped phase $\sigma_{\phi}^2=\sigma_{\phi_w}^2$.

Assuming that the demosaicing algorithm's intensity interpolation is accurate, we have three RGB intensities for every pixel. By \eqref{eqn:phasevariance}, we can find out how uncertain the phase information from each channel is. Then, we can treat it as a \textit{sensor fusion} problem \cite{leon2017probability} with three RGB sensors. Although this assumption seems weak, demosaicing accuracy is increased due to developments of gradient-based \cite{li2008image} and AI-based \cite{gharbi2016deep} demosaicing algorithms, and we believe using the information from a single wavelength sensor is highly susceptible to the second problem mentioned in Section \ref{sect:Introduction}, especially when capturing nonuniform color object.

\section{Proposed Method}
\label{sect:Method}

\noindent In this section, we will introduce our method for solving two problems mentioned in Section \ref{sect:Introduction}. High-level processes of our method, including parameter calibration and 3D reconstruction, are shown in Algorithm \ref{alg:HighlevelProcess}. The hat notation $(\hat{\cdot})$ denotes LCA-corrected images and pixels. Note that there is no hat for the green channel, which is the reference channel. We would like to emphasize the convenience of our calibration setup, which requires only \textbf{two equipments}: checkerboard plate $\mathcal{E}_{c}$ and white plate $\mathcal{E}_{w}$, unlike prior LCA correction algorithms \cite{huang2016analysis, sun2021analysis, chen2020correcting}.

\begin{algorithm}[t]
\caption{LCAMV}\label{alg:HighlevelProcess}
\begin{algorithmic}
\STATE{\textsc{CALIBRATION}}$(\mathcal{E}_c,\mathcal{E}_w)$
  \STATE \hspace{0.5cm}$\mathcal{T} \coloneqq (\mathbf{K}_c,\mathbf{K}_p,\mathbf{R},\mathbf{t}) \gets \text{CalibrateStereo}(\mathcal{E}_{c})$
  \STATE \hspace{0.5cm}$\bm{\theta}_{c} \gets \text{CalibrateCamLCA}(\mathcal{E}_{c})$
  \STATE \hspace{0.5cm}$\bm{\theta}_{p} \gets \text{CalibratePrjLCA}(\mathcal{E}_{w})$
  \STATE \hspace{0.5cm}$\mathbf{k} \gets \text{CalibrateNoise}(\mathcal{E}_{w})$
  \STATE \hspace{0.5cm}\textbf{return} $\mathcal{T},\bm{\theta}_c,\bm{\theta}_p,\mathbf{k}$
\STATE
\STATE{\textsc{RECONSTRUCTION}}$(\mathbf{I},\mathcal{T},\bm{\theta}_c,\bm{\theta}_p,\mathbf{k})$
  \STATE \hspace{0.5cm}$\mathbf{I}\coloneqq(I_1,I_2,\ldots,I_M)$ \COMMENT{Captured $M$ RGB Images}
  \STATE \hspace{0.5cm}$(\mathbf{I}^{R},\mathbf{I}^{G},\mathbf{I}^{B}) \gets \text{ExtractRGB}(\mathbf{I})$
  \STATE \hspace{0.5cm}$(\hat{\mathbf{I}}^{R},{\mathbf{I}}^{G},\hat{\mathbf{I}}^{B}) \gets \text{CorrectCamLCA}(\mathbf{I}^{R},\mathbf{I}^{G},\mathbf{I}^{B},\bm{\theta_c})$
  \STATE \hspace{0.5cm}$(\phi^R,\phi^G,\phi^B) \gets \text{PhaseAcquisition}(\hat{\mathbf{I}}^{R},{\mathbf{I}}^{G},\hat{\mathbf{I}}^{B})$
  \STATE \hspace{0.5cm}$(u_{p}^{R},u_{p}^{G},u_{p}^{B}) \gets \text{ScaleToPixel}(\phi^R,\phi^G,\phi^B)$
  \STATE \hspace{0.5cm}$(\hat{u}_{p}^{R},{u}_{p}^{G},\hat{u}_{p}^{B}) \gets \text{CorrectPrjLCA}(u_{p}^{R},u_{p}^{G},u_{p}^{B},\bm{\theta}_p,\mathcal{T})$
  \STATE \hspace{0.5cm}$\hat{u}_p \gets \text{ChannelFusion}(\hat{u}_{p}^R,{u}_{p}^G,\hat{u}_{p}^B,\hat{\mathbf{I}}^{R},{\mathbf{I}}^{G},\hat{\mathbf{I}}^{B},\mathbf{k})$
  \STATE \hspace{0.5cm}$\hat{z} \gets \text{Triangulate3D}(\mathcal{T},\hat{u}_p)$
  \STATE \hspace{0.5cm}\textbf{return} $\hat{z}$
\end{algorithmic}
\end{algorithm}

\subsection{Lateral Chromatic Aberration Correction}
\label{subsect:LCACorrection}

Regarding the LCA effect, the optical properties of the camera and projector are different across the RGB channel. Thus, we have to consider the LCA effect in the camera and projector stereo-vision calibration process. We suggest using a typical black and white checkerboard instead of a red and blue checkerboard \cite{zhang2006novel}, which could result in inaccurate phase retrieval as mentioned in the first problem in Section \ref{sect:Introduction}. For LCA correction, as shown in \eqref{eqn:LCApolynomial}, we must choose the \textbf{reference channel} to correct other channels. We decided to use the green channel as a reference channel, which has the highest Bayer filter array density and widest overlapping spectrum in visible light. Then, we can get triangulation parameters $\mathcal{T}$ by stereo-vision calibration using only the green channel. We used local homography refinement \cite{moreno2012simple} for better projector calibration quality, despite using a B/W checkerboard.

We follow prior work's camera LCA calibration and correction \cite{mallon2007calibration} method using a B/W checkerboard. Using captured RGB images of a checkerboard, we can extract checkerboard corners from each RGB channel and estimate lateral aberration of each channel with the difference of corner locations. From \eqref{eqn:LCApolynomial}, considering decentering and asymmetry of the image plane, image coordinates are represented as $(x,y)=(au+\dot{u},v+\dot{v})$. Thus, there are total seven parameters $\bm{\theta}_c=\{a,\dot{u},\dot{v},c_1,c_2,c_3,c_4\}$ for LCA modeling in pixel coordinates. By using the green channel as the reference plane, the total camera LCA parameter consists of $\bm{\theta}_c=\{\bm{\theta}_{c}^R,\bm{\theta}_{c}^B\}$ where $R$, $B$ denote the correcting channel. Refer to \cite{mallon2007calibration} for details about the camera LCA model.

\subsubsection{Projector LCA Calibration}

After camera LCA calibration and correction, we can now omit the camera LCA and find out the pure effect of projector LCA. Since we cannot directly get projector pixel $u_p$, we project fringe patterns to a white plate, capture images, conduct phase-shifting \eqref{eqn:arctan_phaseshift}, phase-unwrapping \eqref{eqn:phaseunwrap}, and scale $\phi$ to get $u_p$. Using the camera's RGB channel, we can get projector pixels of different wavelengths $(u_p^R,u_p^G,u_p^B)$. From now on, we will only notate the correction of the red channel for simplicity. The lateral difference at the object point viewed at the camera pixel $(u_c,v_c)$ is:
\begin{equation}
  \Delta_O^R[u_c,v_c]=u_p^G[u_c,v_c]-u_p^R[u_c,v_c]
  \label{eqn:projectorLCA_campixel}
\end{equation}

Recall \eqref{eqn:projector_aberration}; $\Delta_O$ is a linear function of depth, specifically depth from projector $z_p$. Given $u_p^R$, we can triangulate to estimate depth $z_p^R=s_p^R$ and the projector's vertical pixel $v_p^R$ using \eqref{eqn:triangulation_detail} and \eqref{eqn:fundamentalmatrix}. Then, for every camera pixel $(u_c,v_c)$ that is in the projector's Field of View (FoV), we have $(\Delta_O^R,u_p^R,v_p^R,z_p^R)$. Then, we can remap $\Delta_O^R$ and $z_p^R$ into a 2D array in projector pixel $(u_p,v_p)$, and \eqref{eqn:projector_aberration} results in the following:
\begin{equation}
  \Delta_O^R[u_p,v_p]=\alpha^R \cdot z_p^R[u_p,v_p]+\beta^R
\end{equation}

Note that we must use $z_p^R$ calculated from the red channel to correct the red channel independently without using information from other channels. Otherwise, if we use $z_p^G$ or $z_p^B$, noise at other channels will contribute to the red channel during reconstruction. 

Due to the difference between camera and projector resolution, $\Delta_O^R$ will contain empty data points, which can be filled by bilinear interpolation like prior works \cite{mallon2007calibration,huang2016analysis}.

By gathering multiple observations of $\Delta_O^R[u_p,v_p]$ and $z^R[u_p,v_p]$ by varying the white plate's position, we can conduct pixel-wise linear estimation to get 2D arrays of linear coefficients $\alpha^R$ and $\beta^R$. Given two linearly correlated random variables $\Delta_O=\alpha \cdot z_p+\beta$, finding the optimal $\alpha^*$ and $\beta^*$ using the linear minimum mean square error condition \cite{kay1993fundamentals,leon2017probability} is as follows:
\begin{subequations}
\begin{align}
  \alpha^*&=\frac{Cov[\Delta_O,z_p]}{Var[z_p]}=\rho_{\Delta_O,z_p}\frac{\sigma_{\Delta_O}}{\sigma_{z_p}}\\
  \beta^*&=E[\Delta_O]
\end{align}
\end{subequations}

\begin{figure}[t]
  \centering
  \includegraphics[width=88mm]{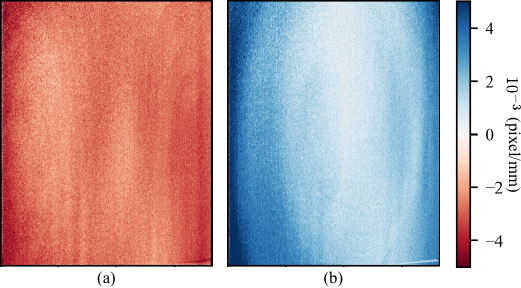}
  \includegraphics[width=88mm]{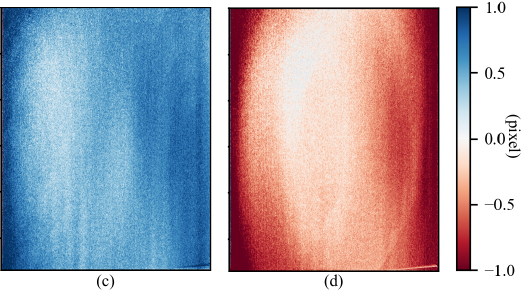}
  \caption{Projector LCA coefficient map (a) $\alpha^R$ (b) $\alpha^B$ (c) $\beta^R$ (d) $\beta^B$}
  \label{fig:alphamap}
\end{figure}

Using sample mean and sample variance, we can estimate $\alpha^R$ and $\beta^R$ using data points of $\Delta_O^R$ and $z_p^R$. The blue channel can be corrected in the same manner, resulting in a total of four 2D maps $\bm{\theta}_p=\{\alpha^R,\beta^R,\alpha^B,\beta^B\}$ for projector LCA correction, as shown in Fig. \ref{fig:alphamap}. Note that the unit of $\alpha$ is [pixel/mm], lateral aberration in pixel per depth. 

We adjusted white plate locations gradually, increasing $z_p$ from 180 to 340 millimeters far from the projector to cover the camera-projector FoV range, and we used a total of $N=18$ white plate locations for samples.

\subsubsection{Camera Projector LCA correction}

Assuming we now have all parameters $\{\bm{\theta}_c,\bm{\theta}_p\}$, we can correct RGB images and projector pixels. For image $I^R$, we can calculate the corrected pixel and bilinearly interpolate to reconstruct the image without the aberration:
\begin{subequations}
\begin{align}
  (\hat{u}_c^R,\hat{v}_c^R&)=(u_c+\Delta_x^R(u_c,v_c,\bm{\theta}_c^R),v_c+\Delta_y^R(u_c,v_c,\bm{\theta}_c^R)) \\ 
  \hat{I}^R&=\text{Interpolate}(I^R,\hat{u}_c^R,\hat{v}_c^R)
\end{align}
\end{subequations}

By camera LCA corrected images $\hat{\mathbf{I}}^R$, we can conduct phase shifting and phase unwrapping to find out projector pixel $u_p^R$. Then, projector LCA correction is done as follows:
\begin{subequations}\label{eqn:Project-LCA-Correct}
\begin{align}
  (z_p^R, v_p^R) &= \text{Triangulate}(u_p^R, \mathcal{T}) \\
  \hat{u}_p^R &= u_p^R + \Delta_O^R(u_p^R,v_p^R,z_p^R,\alpha^R,\beta^R)
\end{align}
\end{subequations}

where $\text{Triangulate}(\cdot)$ is done by solving \eqref{eqn:triangulation_detail} for $z_p^R$ and \eqref{eqn:fundamentalmatrix} for $v_p^R$. For most cases where $(u_p,v_p)$ are not integers, we can bilinear interpolate with the given $\alpha$ and $\beta$ map. Note that the projector LCA correction of a certain channel is done independently using information from that channel, and the correction is done only for the horizontal pixel, unlike both pixels in the camera LCA correction.

\subsection{Minimum Variance Channel Fusion}
\label{subsect:RGBFusion}

Suppose we now have properly aligned projector pixels $(\hat{u}_p^R,u_p^G,\hat{u}_p^B)$ from each channel. We can merge information using the \textit{Minimum Variance Unbiased} (MVU) estimator, where the variance can be calculated as \eqref{eqn:phasevariance}. To find out the noise level for each channel, we first have to calibrate for noise coefficients $k_0$ and $k_1$.

\subsubsection{Intensity Noise Calibration}

Our intensity noise calibration adopts simple procedures \cite{healey2002radiometric} using sample mean and sample variance with varying light conditions. Since we can control the light source by manipulating the projector, we easily apply the prior approach into DFP. Upon projecting $P$ different uniform intensities to the white plate, capture two images for each. For each pair $(I^u_1,I^u_2)$, the difference between intensity is zero mean $I^u_1-I^u_2=I^u_\Delta\sim \mathcal{N}(0,2\sigma_n^2)$. We use pixel-wise sample mean and sample variance to estimate noiseless intensity ${\mu_I}$ and noise $\sigma_n^2$:
\begin{subequations}
\begin{align}
  \hat\mu_I&=\frac{1}{2|\mathcal{D}|}\sum_{(u_c,v_c)\in\mathcal{D}}{I^u_1[u_c,v_c]+I^u_2[u_c,v_c]}\\
  \hat\mu_\Delta&=\frac{1}{|\mathcal{D}|}\sum_{(u_c,v_c)\in\mathcal{D}}{I^u_1[u_c,v_c]-I^u_2[u_c,v_c]}\\
  \hat{\sigma}_n^2&=\frac{1}{2(|\mathcal{D}|-1)}\sum_{(u_c,v_c)\in\mathcal{D}}{(I^u_\Delta[u_c,v_c]-\hat{\mu}_\Delta)^2}
\end{align}
\end{subequations}

where $\mathcal{D}$ denotes the pixel domain where nearly the same level of light is projected, $\hat{\mu}_\Delta$ is the domain sample mean of $I_\Delta$, and $\hat{\sigma}_n$ is the estimated sample variance. Given $P$ data points of $(\hat{\mu}_I,\hat{\sigma}_n^2)$ pairs, we can conduct weighted line fitting, which is a maximum likelihood estimate for $(k_0,k_1)$:
\begin{equation}
  \chi^2=\sum_{i=1}^P{\frac{(\hat{\sigma}_{n_i}^2-(k_0+k_1\cdot \hat{\mu}_{I_i}))^2}{Var[\hat{\sigma}_{n_i}^2]}}
  \label{eqn:noiseMLE}
\end{equation}

where $Var[\hat{\sigma}_n^2]$ can be approximated under the Gaussian assumption:
\begin{equation}
  Var[\hat{\sigma}_n^2]\approx \frac{2(\hat{\sigma}_n^2)^2}{(|\mathcal{D}|-1)}
\end{equation}

\begin{figure}[t]
    \centering
    \includegraphics[width=88mm]{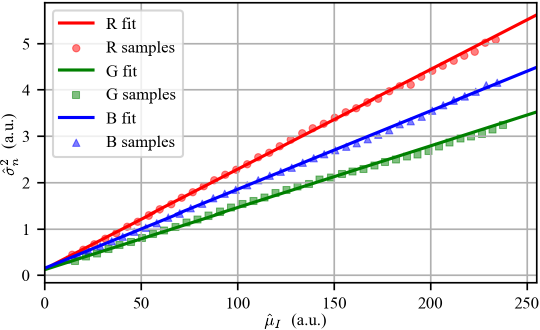}
    \caption{Calibrated RGB Noise Model Fit}
    \label{fig:RGBnoise}
\end{figure}

We can estimate noise parameters $(k_0,k_1)$ by minimizing \eqref{eqn:noiseMLE}. Refer to \cite{healey2002radiometric} for details about the noise calibration model. By calibrating for all channels, we have total six parameters. $\mathbf{k}=\{k_0^R,k_1^R,k_0^G,k_1^G,k_0^B,k_1^B\}$. Fig. \ref{fig:RGBnoise} and Table. \ref{tab:noisecoeff} is the calibrated result for the Poisson-Gaussian noise model, using $P=40$ different intensity levels with $|D|\approx140000$ pixels. It can be found out that the red channel is the noisiest channel and the green channel is the least. 

\begin{table}[t]
\begin{center}
\caption{Calibrated RGB Noise Coefficient}
\label{tab:noisecoeff}
\begin{tabular}{|c|c|c|c|}
\hline
 & Red & Green & Blue\\
\hline
$k_0$ & 0.1333 & 0.1184 & 0.1500 \\
$k_1$ & 0.0215 & 0.0134 & 0.0170 \\
\hline
\end{tabular}
\end{center}
\end{table}

\subsubsection{RGB Channel Fusion}

Given noise parameters $\mathbf{k}$, we can find out the variance of the phase from each channel using \eqref{eqn:phasevariance_PoissonGaussian}. We use the weighted average to estimate the optimal $\hat{u}_p$, where the following should hold:
\begin{subequations}
\begin{align}
    \hat{u}_p&=w_R\hat{u}_p^R+w_Gu_p^G+w_B\hat{u}_p^B\\
   1&=w_R+w_G+w_B \label{eqn:weight_unbiased}\\
   \mathbf{w}&=\arg\min_{\mathbf{w}}(Var[\hat{u}_p]) \label{eqn:weight_minvar}
\end{align}
\end{subequations}

where $\mathbf{w}=[w_R,w_G,w_B]$ is the weight vector, \eqref{eqn:weight_unbiased} denotes the unbiased condition and \eqref{eqn:weight_minvar} denotes the minimum variance condition. Using the Poisson-Gaussian assumption, where each channel is statistically independent, variance leads to:

\begin{equation}
    Var[\hat{u}_p]=w_R^2Var[\hat{u}_p^R]+w_G^2Var[u_p^G]+w_B^2Var[\hat{u}_p^B]
\end{equation}

The variance of the projector pixel is simply a scaled version of the variance of the phase. Neglecting the uncertainty induced by the LCA correction, we can calculate the uncertainty of the projector pixel as

\begin{subequations}\label{eqn:MVUcondition}
\begin{align}
  Var[\hat{u}_p]&=w_R^2\sigma_{u_p^R}^2+w_G^2\sigma_{u_p^G}^2+w_B^2\sigma_{u_p^B}^2 \\
  \sigma_{u_p^i}^2&= (\lambda/2\pi)^2\cdot \sigma_{\phi^i}^2, \quad i\in\{R,G,B\}
\end{align}
\end{subequations}

where $(\lambda/2\pi)$ is the coefficient scaling phase to projector pixel, and $\sigma_{\phi^i}^2$ is the uncertainty of phase from each channel calculated by \eqref{eqn:phasevariance_PoissonGaussian}. The solution for minimizing \eqref{eqn:MVUcondition} with constraints can be done using the Cauchy-Schwartz inequality. Let two vectors be $\mathbf{a}_1=[w_R\sigma_{u_p^R},w_G\sigma_{u_p^G},w_B\sigma_{u_p^B}], \mathbf{a}_2=[1/\sigma_{u_p^R},1/\sigma_{u_p^G},1/\sigma_{u_p^B}]$. Dividing $\|\mathbf{a_2}\|^2$ to the each side of $|\langle\mathbf{a}_1,\mathbf{a}_2\rangle|^2\leq \|\mathbf{a}_1\|^2\|\mathbf{a}_2\|^2$ leads to:
\begin{equation}
\begin{aligned}
  \frac{(w_R+w_G+w_B)^2}{\|\mathbf{a}_2\|^2}=\frac{1}{{\|\mathbf{a}_2\|^2}}\leq (w_R^2\sigma_{u_p^R}^2+w_G^2\sigma_{u_p^G}^2+w_B^2\sigma_{u_p^B}^2)
\end{aligned}
\end{equation}

Then the right side $Var[\hat{u}_p]$ is lower bounded, and the equality holds when $\mathbf{a}_1$ and $\mathbf{a}_2$ are the scalar multiple of each other. Under this condition, solving for optimal $\mathbf{w}^*$ leads to:
\begin{equation}
  w_i^*=\frac{1/\sigma_{u_p^i}^2}{\sum_{j}{1/\sigma_{u_p^j}^2}},\quad i,j\in\{R,G,B\}
\label{eqn:optimalweight}
\end{equation}

Note that optimal scalar weights $\mathbf{w}^*$ can be calculated deterministically after finding uncertainties. Since $w_R,w_G,w_B$ are never 0 unless $\sigma_{u_p^i}^2$ goes to infinity, this method is still susceptible to the outlier information from noisy channels. Since the phase unwrapping algorithm incorporates discontinuities, the outlier channel can have catastrophic $2\pi$ jump errors. To remove the effect of the outlier channel, we exploit the fact that we know the variance $\sigma_{u_p^i}^2$ of the projector pixel, although its distribution is unknown and non-Gaussian.

\begin{figure}[t]
  \centering
  \includegraphics[width=88mm]{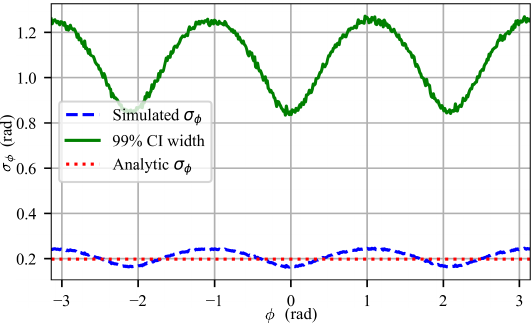}
  \caption{Standard deviation and confidence interval of $\phi$}
  \label{fig:montecarlo}
\end{figure}

We use the \textit{Monte Carlo method} \cite{metropolis1949monte} to approximate the Confidence Interval (CI) of the unknown $\phi$ distribution. Note that $u_p$ is simply the scalar multiple of $\phi$; thus, the CI of $u_p$ is also the scaled version of $\phi$. For the Monte Carlo method, we used a total of $N=10000$ samples with the feasible intensity noise level $k_0=0.0133, k_1=0.1212$ from the noise calibration of the Poisson-Gaussian model \eqref{eqn:poissongaussian}, and the feasible ratio between mean and modulation intensities ($I_A:I_B=2:1$). We simulated three intensities using \eqref{eqn:3stepshift} and forwarded them to the three-step phase shifting algorithm \eqref{eqn:arctan_phaseshift} to approximate the unknown distribution of $\phi$, ranging in $[-\pi,+\pi)$. Note that the phase-shifting algorithm returns wrapped phase; thus, circular mean and circular standard deviation \cite{mardia2009directional} were used to calculate statistics:
\begin{subequations}
\begin{align}
  \bar{R} &= \frac{1}{N} \sum_{i=1}^N \exp(j \phi_i), \\ 
  \mu_{\phi_{2\pi}} &= \arg(\bar{R}), \\
  \sigma_{\phi_{2\pi}} &= \sqrt{-2 \ln |\bar{R}|}
\end{align}
\end{subequations}

where $\mu_{\phi_{2\pi}}$ is the circular mean, $ \sigma_{\phi_{2\pi}}$ is the circular standard deviation, and $j$ is the imaginary unit. As shown in Fig. \ref{fig:montecarlo}, it could be found that the simulated sample standard deviation oscillates around the analytically calculated $\sigma_\phi$ using \eqref{eqn:phasevariance_PoissonGaussian} in the phase domain. The mean $99\%$ CI across the phase domain was found to be $\pm 2.72\sigma_\phi$.

To filter out the outlier channel, we define the least uncertain channel as the anchor channel and find if the $u_p$ from other channels are in the anchor channel's $99\%$ CI. If the projector pixel from a certain channel is not in the CI, we mask that channel's uncertainty to infinity and reassign the weights using \eqref{eqn:optimalweight}. The outlier channel filtering process can be mathematically expressed as:
\begin{subequations}\label{eqn:weight_filtering}
\begin{align}
  \sigma_{u_p^i}'^2 &= 
  \begin{cases}
    \infty, & \text{if } \left| u_p^i - u_p^{\dagger} \right| > 2.72 \, \sigma_{u_p^{\dagger}} \\
    \sigma_{u_p^i}^2, & \text{otherwise}
  \end{cases} \\
    \dagger &\coloneqq \underset{i}{\arg\min} (\sigma_{u_p^i}^2), \quad i \in \{R, G, B\}
\end{align}
\end{subequations}

where $\dagger$ is the anchor channel and $\sigma_{u_p^i}'^2$ is the refined uncertainties to re-assign the weights, neglecting the outlier channel. Note that the hat notation is skipped, but $u_p^i, u_p^\dagger$ denotes the LCA-corrected projector pixel if the channel is red or blue. Then, the reassigned weights will omit the effect of the outlier channel. Given new weights, the optimal $\hat{u}_p$ can be calculated, and the 3D geometry $\hat{z}$ can be acquired by triangulation \eqref{eqn:triangulation_detail}. 

\begin{figure}[!t]
  \centering
  \includegraphics[width=88mm]{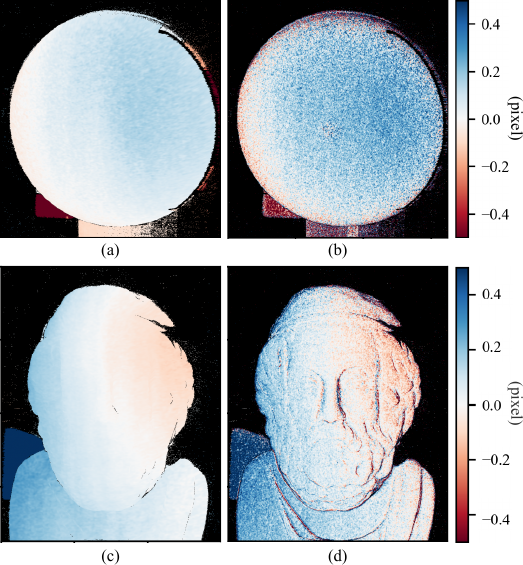}
  \caption{$\hat{\Delta}_O, \Delta_O$ while scanning white sphere and sculpture (a) $\hat{\Delta}_O^R$ (b) $\Delta_O^R$ (c) $\hat{\Delta}_O^B$ (d) $\Delta_O^B$}
  \label{fig:diffPlot}
\end{figure}
\begin{figure}[!t]
  \centering
  \includegraphics[width=0.38\textwidth]{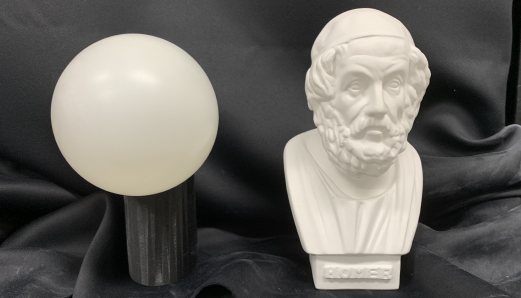}
  \caption{Image of White Sphere and Sculpture}
  \label{fig:sphere_sculpture}
\end{figure}

\section{Experimental Result}
\label{sect:Experiment}

\noindent In this section, we will show the validity of LCAMV by real-world experiments. In Section \ref{subsect:quantitative}, we will show experimental results explaining one of our main contributions: the projector LCA model and correction. Then, we will show increased 3D geometry precision of LCAMV compared to typical RGB to grayscale intensity baselines. In Section \ref{subsect:ablation}, we will show the LCA correction stage and the minimum variance channel fusion stage effect on precise 3D reconstruction of colored objects by ablation study. Lastly, we will compare the quality of 3D geometry of nonuniform color objects with complex geometry using the proposed method with other baselines. All of our experiments were conducted with the Grasshopper3 GS3-U3-23S6C-C and TI DLP LightCrafter 4500 for hardware setups.

\subsection{Quantitative Analysis}

\label{subsect:quantitative}

\subsubsection{LCA Model Analysis} We have weakly proven the validity of our theoretical model for LCA at the object $\Delta_O$ with scanning white plates and using the Pearson correlation coefficient in Section \ref{sect:Preliminary}. However, can this model be applied to general non-planar objects? If our theoretical model is truly valid, the estimated $\hat{\Delta}_O$ and the observed $\Delta_O$ should show the similar trend on objects of any geometry. To find out, we calculated the observed lateral aberration $\Delta_O^R[u_c,v_c]=u_p^G[u_c,v_c]-u_p^{R}[u_c,v_c]$, while scanning the white non-planar objects. We compare $\hat{\Delta}_O$ and $\Delta_O$ by scanning the white, complex objects as shown in Fig. \ref{fig:diffPlot}. Two objects are shown in Fig. \ref{fig:sphere_sculpture}. The top two figures are $\Delta_O^B$ and $\hat{\Delta}_O^B$ while scanning the white sphere, and the two figures at the bottom are $\Delta_O^R$ and $\hat{\Delta}_O^R$ while scanning the white sculpture. It can be found out that the projector LCA of the red channel is about +0.3 pixel, and the blue channel varies around -0.25 pixel to +0.25 pixel on each object. Comparing the actual and the estimated projector LCA, $\Delta_O$ looks like a noisy version of $\hat{\Delta}_O$ due to stochasticity, but the overall heatmap across local regions looks similar.

\begin{figure}[t]
  \centering
  \includegraphics[width=88mm]{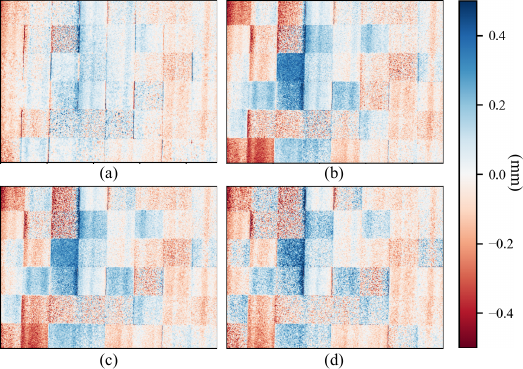}
  \caption{Colorboard plane fitting error map using (a) LCAMV (b) Mean (c) Y'UV (d) Green}
  \label{fig:colorboard_errormap}
\end{figure}

\begin{figure}[t]
  \centering
  \includegraphics[width=44mm]{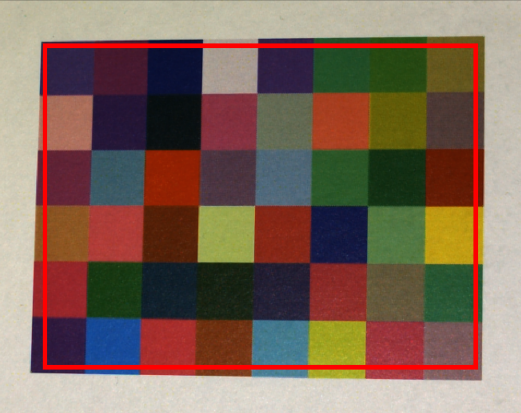}
  \caption{ROI and image of Colorboard}
  \label{fig:colorboard}
\end{figure}

\subsubsection{Colorboard Plane Fitting} To prove the improved precision of LCAMV in scanning nonuniform color objects, we have created a colorboard with checker patterns of a total of $6\times8$ random colors. Then, we have printed out the colorful checker pattern and attached it to the flat plate, as flat as possible. To show the LCAMV's superiority across all general phase-shifting algorithms, we scanned a colorboard using 3, 12, and 18-step phase-shifted patterns with a wavelength of $\lambda=36$ pixels and a phase period of $L=32$ using binary graycode decoding for the phase unwrapping algorithm, with an exposure time of 7 milliseconds. To minimize the camera and projector distortion effect on plane fitting error, we placed the colorboard plate far enough away, about $z\approx320$ millimeters away from the camera. From the acquired depth image, we manually masked the Region of Interest (ROI) of the colorboard as shown in Fig. \ref{fig:colorboard}. Since the number of point clouds in ROI is too large $(N\approx160000)$ for the plane fitting algorithm, we randomly subsampled 10,000 points in ROI to define the plane and used all points to calculate the plane fitting error. We compared LCAMV with other baselines using mean intensity, Y'UV transform, and only the green channel. 

\begin{table}[t]
\begin{center}
\caption{Colorboard Plane Fitting MSE}
\label{tab:ColorboardPlaneFit}
\begin{tabular}{|c|c|c|c|c|}
\hline
Method & \multicolumn{4}{c|}{MSE ($\text{mm}^2$)} \\
\cline{2-5}
 & LCAMV & Mean & Y'UV & Green\\
\hline
3-Step & \textbf{0.036073} & 0.055645 & 0.064706 & 0.082067 \\
12-Step & \textbf{0.014794} & 0.027995 & 0.029057 & 0.038311 \\
18-Step & \textbf{0.012485} & 0.024468 & 0.024343 & 0.032854 \\
\hline
\end{tabular}
\end{center}
\end{table}

For all baselines, we could find out that an increased number of patterns leads to better precision, which is theoretically correct. Most importantly, we could find out that LCAMV's plane fitting error is consistently lower than any other baseline. Compared to the second-best baselines, LCAMV shows $\textbf{43.6\%}$ reduced plane fitting error on average, which shows the robust performance of LCAMV. Fig. \ref{fig:colorboard_errormap} is a comparison of the error map of each baseline where 18 steps were used. It can be found out that Mean, Y'UV, and Green exhibit error regions clearly divided into blocks of yellow, cyan, red, and blue colors, which is due to the LCA effect induced by different colors of checker blocks. While LCAMV's error is similar locally across checker blocks, despite the small portion of yellow and cyan regions due to camera and projector distortion. For all baselines, it could be seen that high error is present at the vertical block boundaries, which are known as Discontinuity-induced Measurement Artifacts (DMA) \cite{yue2019reduction, blanchard2022removal, chen2017analysis}. DMA is induced by the blur at the abrupt change in irradiation level, rather than irradiation wavelength. Since DMA is unrelated to the LCA effect, we will treat DMA correction as out of this paper's scope. The green channel's performance was the worst, due to the absence of the averaging effect of noise, unlike Mean and Y'UV. Surprisingly, using only the green channel could not solve the LCA problem perfectly, as shown in Fig. \ref{fig:colorboard_errormap}. This is because both the projector's and the camera's channels treat color as wavelengths of a continuous spectrum, not a single wavelength. Since the projector and the camera do not project and scan exactly one single wavelength, overlapping spectra of red and blue lights still affect the information in the green channel. Inferred from this result, using a monochrome camera can also suffer from LCA when scanning nonuniform colored objects; thus, LCAMV is the only promising solution when scanning nonuniform color objects. We will also qualitatively show that using the green channel cannot fully solve the LCA effect in Section \ref{subsect:qualitative}.

\begin{figure*}[t]
  \centering
  \subfloat[]{\includegraphics[width=58mm]{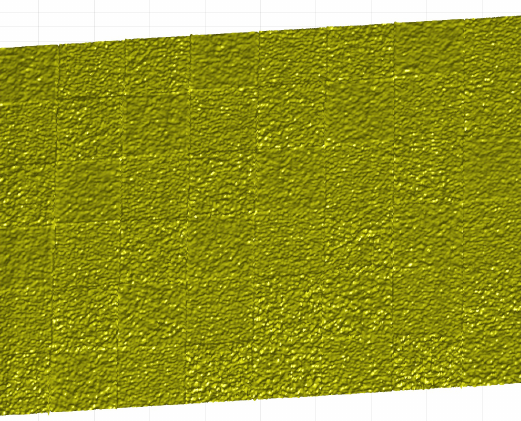}}\hfill
  \subfloat[]{\includegraphics[width=58mm]{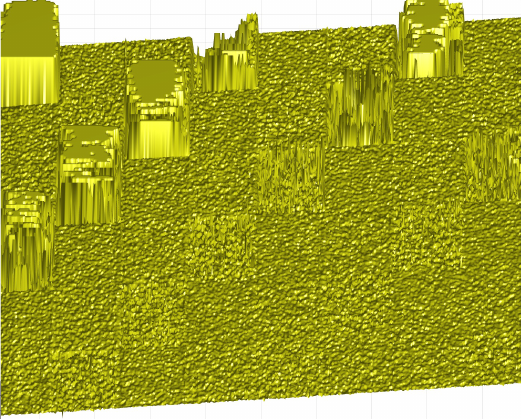}}\hfill
  \subfloat[]{\includegraphics[width=58mm]{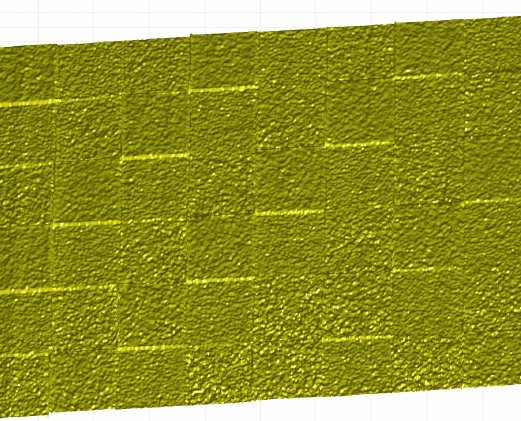}}
  \caption{RGB board rendering comparison using (a) LCAMV (b) LCA (c) MV}
  \label{fig:RGB_board_render}
\end{figure*}

\subsection{Ablation Study}

\label{subsect:ablation}

\begin{figure}[t]
  \centering
  \includegraphics[width=44mm]{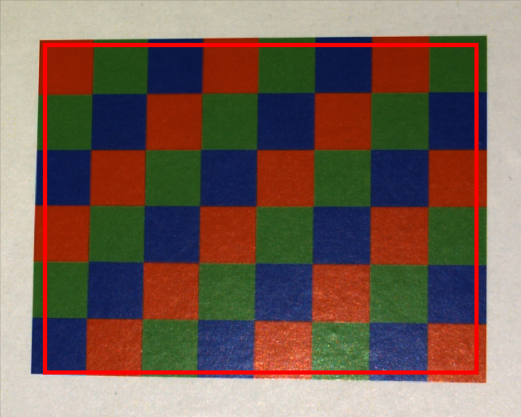}
  \caption{ROI and image of colorboard}
  \label{fig:rgbboard}
\end{figure}

Next, for further understanding of LCAMV's stages, we conduct the ablation study with two modified methods: LCA and MV. LCA only conducts LCA correction, but the resulting projector pixel is the mean of LCA corrected projector pixels from each channel. MV does not conduct any LCA corrections, but projector pixels from each channel are merged with the MVU manner. To clearly visualize the difference and observe each stage's contribution, we created an RGB board, similar to the prior experiment. We compare three methods with plane fitting error and rendered visualization of 18-step reconstruction. Experiment setups are the same as with the prior colorboard plane fitting experiment, and the image of the RGB board and ROI is shown in Fig. \ref{fig:rgbboard}.

Plane fitting mean squared error was the lowest at LCAMV for all methods. Note that the LCA method failed and showed extremely high plane fitting error due to the absence of weight filtering described in \eqref{eqn:weight_filtering}. Fig. \ref{fig:RGB_board_render} is the reconstructed rendering of the RGB board reconstructed using the 18-step phase-shifting algorithm. It can be found out that block artifacts are visible in MV unlike LCAMV. Although LCA fixes the block artifacts by LCA correction, the reconstructed surface looks comparatively rough due to high noise, and the blocks of red color at the top left region are corrupted by the outlier blue channel's catastrophic $2\pi$ jump error. On the other hand, the incorporated method LCAMV corrects LCA effectively, exhibits less noise using the MVU condition, and omits the effect of the outlier channel. The ablation study shows that both the LCA correction stage and the minimum variance fusion stage work complementarily to achieve the best performance.

\begin{table}[t]
\begin{center}
\caption{RGB board Plane Fitting MSE}
\label{tab:RGBplanefit}
\begin{tabular}{|c|c|c|c|}
\hline
Method & \multicolumn{3}{c|}{MSE ($\text{mm}^2$)} \\
\cline{2-4}
       & LCAMV & LCA (Abl.) & MV (Abl.)\\
\hline
3-Step  & \textbf{0.030582} & 282.76 & 0.066288 \\
12-Step & \textbf{0.016121} & 295.12 & 0.054010 \\
18-Step & \textbf{0.014217} & 299.32 & 0.052371 \\
\hline
\end{tabular}
\end{center}
\end{table}

\subsection{Qualitative Analysis}

\label{subsect:qualitative}

\begin{figure*}[htbp]
  \centering
  \subfloat[]{\includegraphics[width=58mm]{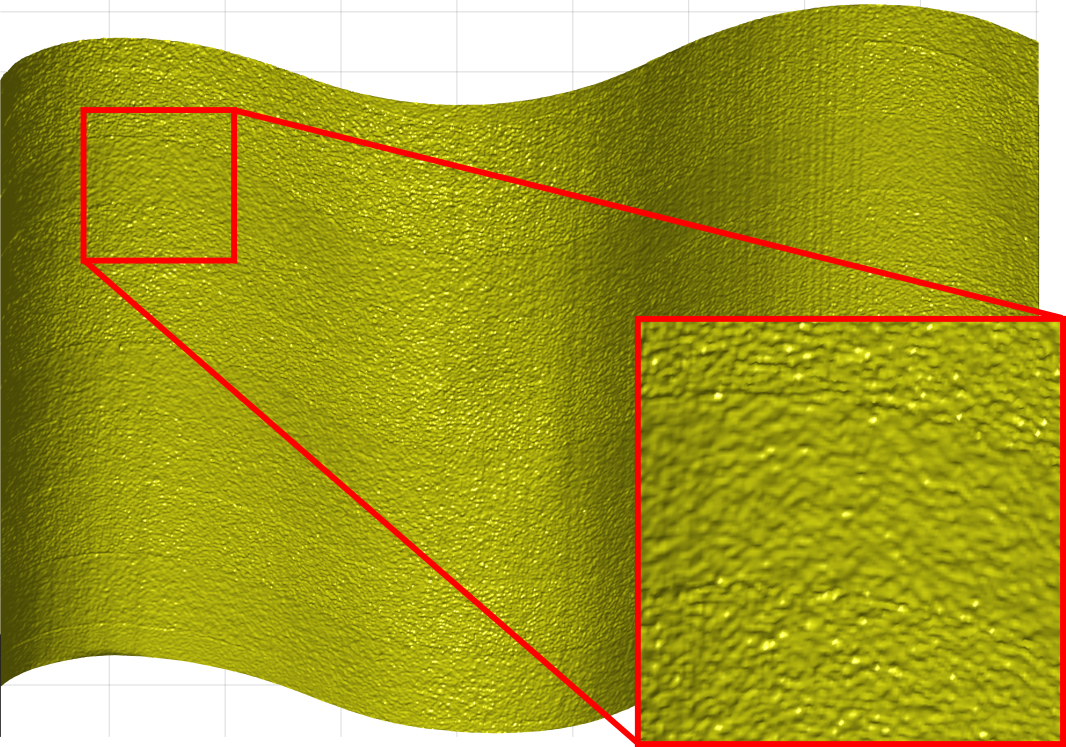}}\hfill
  \subfloat[]{
  \includegraphics[width=58mm]{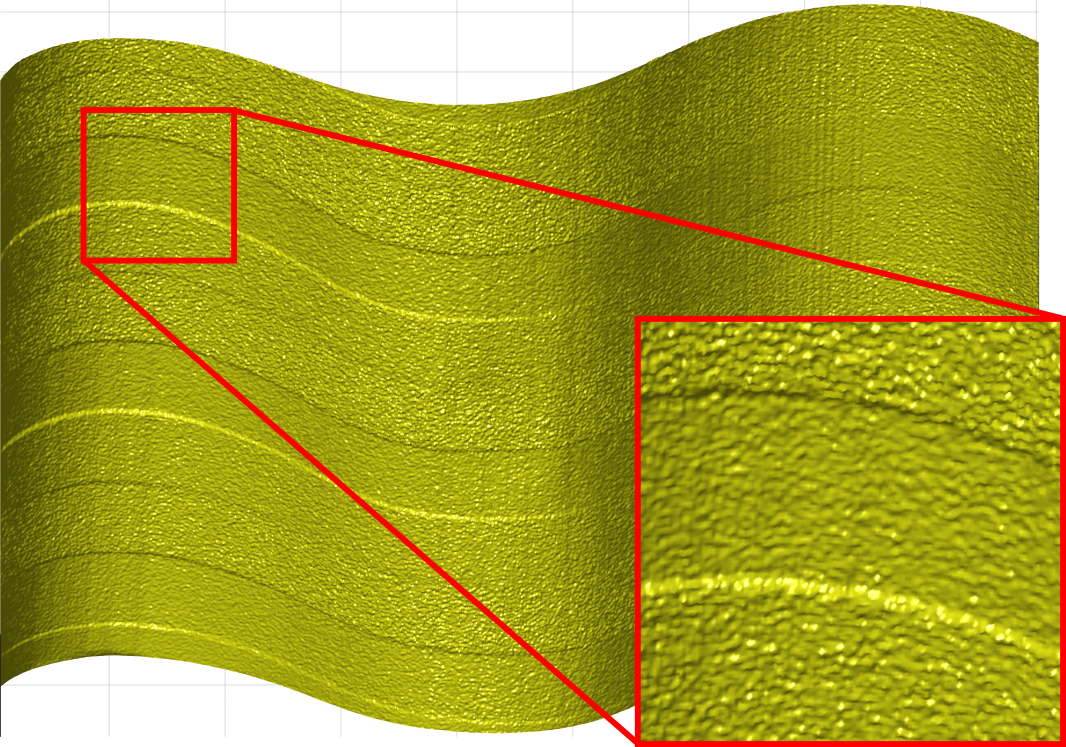}}\hfill
  \subfloat[]{
  \includegraphics[width=58mm]{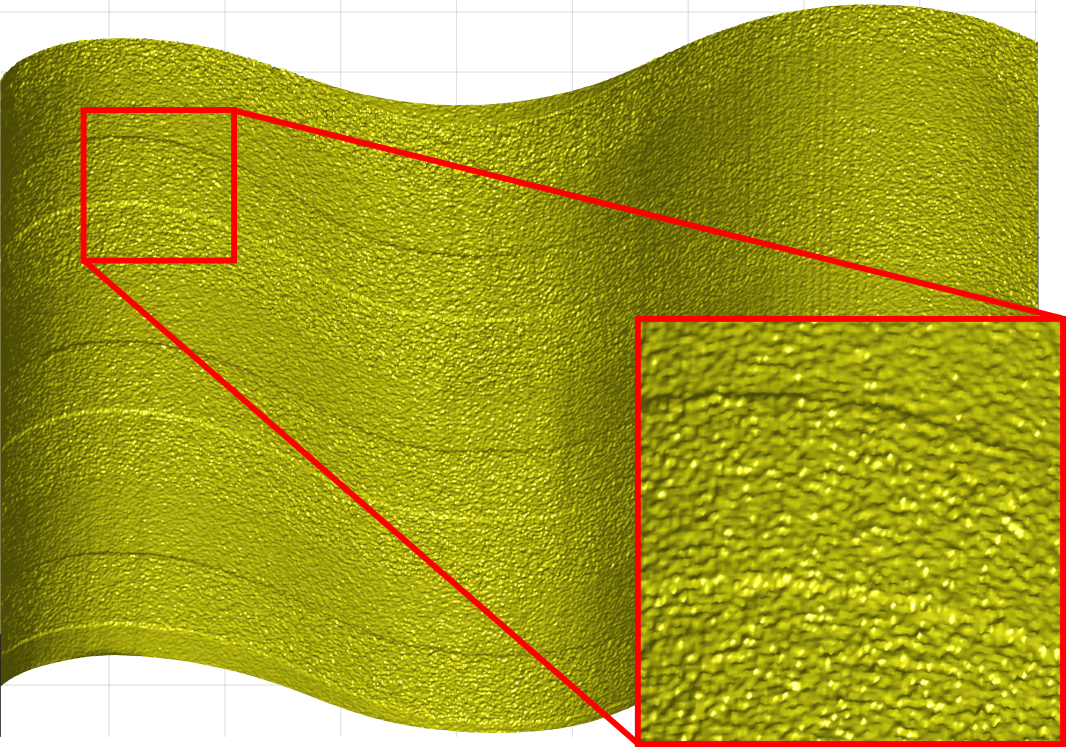}}
  \caption{Curvy RGB surface rendering comparison using (a) LCAMV (b) Mean (c) Green}
  \label{fig:qualitative_render_curve}
\end{figure*}

\begin{figure*}[t]
  \centering
  \subfloat[]{  \includegraphics[width=58mm]{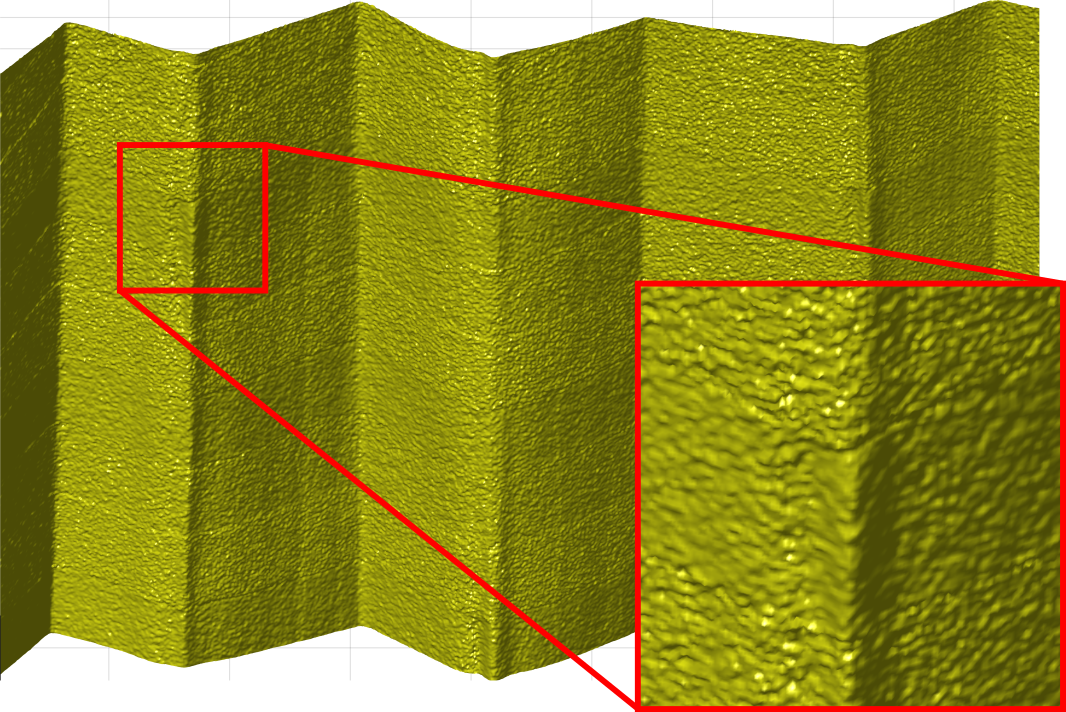}}\hfill
  \subfloat[]{
  \includegraphics[width=58mm]{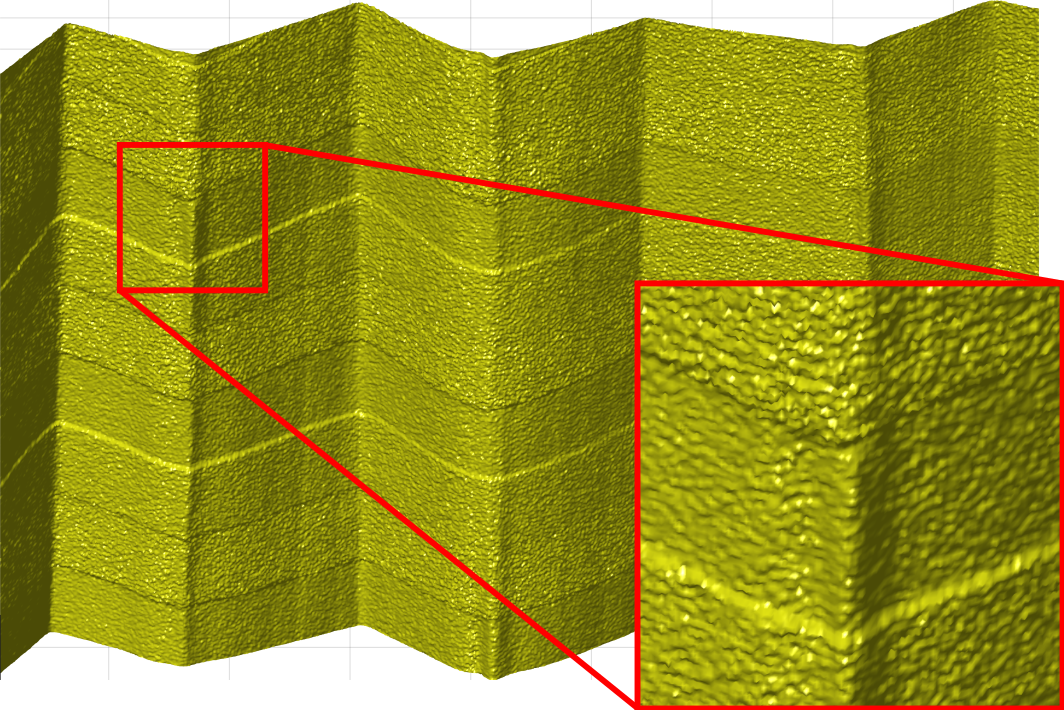}}\hfill
  \subfloat[]{
  \includegraphics[width=58mm]{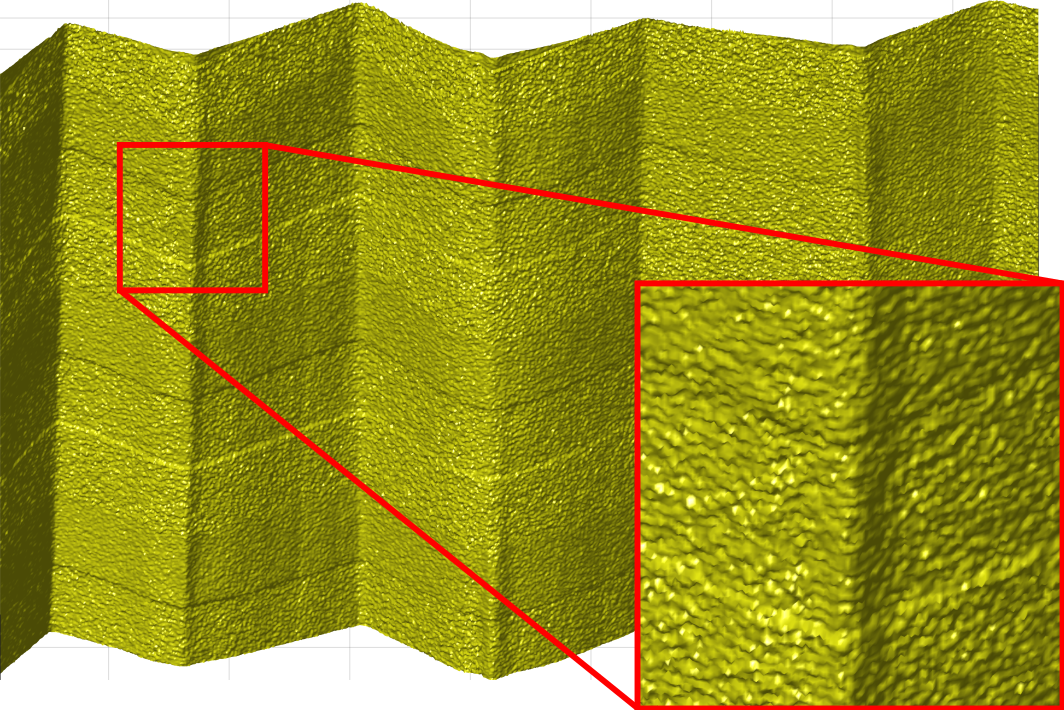}}
  \caption{Zigzagged RGB surface rendering comparison using (a) LCAMV (b) Mean (c) Green}
  \label{fig:qualitative_render_zigzag}
\end{figure*}

\begin{figure}[t]
  \centering
  \includegraphics[width=88mm]{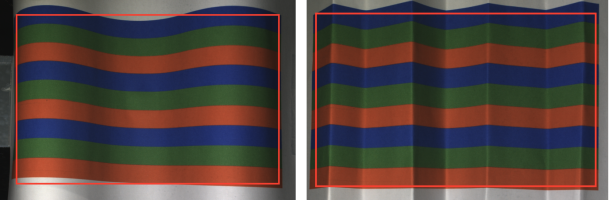}
  \caption{ROI and image of RGB surface (a) curvy (b) zigzagged}
  \label{fig:qualitative_surface}
\end{figure}

Lastly, to extensively validate LCAMV's performance on general non-planar nonuniform color objects, we printed out RGB stripe patterned paper and arbitrarily created a curvy surface and a zigzagged surface as shown in Fig. \ref{fig:qualitative_surface}. We compare the ROI region's rendered result with the second-best baseline from colorboard plane fitting: mean intensity and the green channel to test if the pseudo-monochrome camera can solve the LCA effect. We used the 18-step phase-shifting algorithm and 6 milliseconds for the exposure time for qualitative analysis. The remaining experiment setups are the same as the colorboard plane fitting experiment.

As shown in Fig. \ref{fig:qualitative_render_curve} and Fig. \ref{fig:qualitative_render_zigzag}, it is clearly visible that rendered rows of different colors show different surface levels for the mean and green baselines due to the LCA effect. Unlike Mean and Green, the rendered result of LCAMV does not show any bumpy artifacts, and it is hard to tell where the color stripes' boundaries are located. Surprisingly, using the pseudo-monochrome information from the green channel showed less bumpy artifacts than the mean intensity, but it could not fully solve the LCA effect. Again, this result shows that LCAMV with the RGB camera is the only solution to solve the LCA effect while scanning nonuniform color objects.

\section{Conclusion}
\label{sect:Conclusion}

\noindent This paper proposes LCAMV, the algorithm that can robustly reconstruct 3D geometry of nonuniform color objects. LCAMV consists of two stages: LCA correction and minimum variance channel fusion. The LCA correction stage effectively compensates for the bias of each channel, especially that induced by the projector's LCA. The minimum variance channel fusion stage merges the information from each channel with the highest precision. We believe that LCAMV is the solution bridging DFP in controlled laboratory conditions to more practical and industrial conditions, where the objects could be in any color, and accurate RGB-D data is required. 

However, the complexity of the algorithm is a limitation for future improvements. Although every part of the LCAMV algorithm is deterministic and does not include any iterative optimizations, the computation burden is obviously larger than the typical phase-shifting and triangulation algorithm. Optimization of the LCAMV algorithm might be required for the real-time RGB-D acquisition.

\section*{Acknowledgement}
This research was supported by the Technology Innovation Program(Project Name: Development of AI autonomous continuous production system technology for gas turbine blade maintenance and regeneration for power generation, Project Number: RS-2025-25447257, Contribution Rate: 50\%) funded By the Ministry of Trade, Industry and Resources(MOTIR, Korea), the Culture, Sports and Tourism R\&D Program through the Korea Creative Content Agency grant funded by the Ministry of Culture, Sports and Tourism in 2024 (Project Name: Global Talent for Generative AI Copyright Infringement and Copyright Theft, Project Number: RS-2024-00398413, Contribution Rate: 40\%), and the National Research Foundation of Korea(NRF) grant funded by the Korea government(MSIT) (Project Number: RS-2025-16072782, Contribution Rate: 10\%).

\bibliographystyle{IEEEtran}
\bibliography{ref}

%\vspace{-22pt}
%\begin{IEEEbiography}[{\includegraphics[width=1in,height=1.25in,clip,keepaspectratio]{fig/Photo_OWB.png}}]{Wonbeen Oh}
%   received the B.S. degree in mechanical engineering from Yonsei University, Seoul, South Korea. He is currently working toward the M.S. degree in mechanical engineering at Yonsei University. His current research interests include 3D optical sensing, machine vision and deep learning. 
%\end{IEEEbiography}

%\vspace{-22pt}
%\begin{IEEEbiography}
%[{\includegraphics[width=1in,height=1.25in,clip,keepaspectratio]{fig/Photo_HJS.png}}]{Jae-Sang Hyun}
%   is an assistant professor in the Department of Mechanical Engineering, Yonsei University, Seoul, South Korea. He worked at ORBBEC 3D, Troy, MI, USA as a research scientist for the development of 3D cameras. He received his Ph.D. in mechanical engineering from Purdue University and B.S. degree in mechanical engineering from Yonsei University.  His major research areas include 3D optical sensing, 3D reconstruction, and 3D sensor development.
%\end{IEEEbiography}

%\vfill

\end{document}